\documentclass[letterpaper]{article} 
\usepackage{aaai2026}  
\usepackage{times}  
\usepackage{helvet}  
\usepackage{courier}  
\usepackage[hyphens]{url}  
\usepackage{graphicx} 
\usepackage{natbib}  
\usepackage{caption} 
\usepackage{algorithm}
\usepackage{algorithmic}

\usepackage{booktabs}

\usepackage{amssymb} 

\usepackage{amsmath} 
\newcommand{\argmin}{\mathop{\mathrm{argmin}}}

\usepackage{newfloat}
\usepackage{listings}
\DeclareCaptionStyle{ruled}{labelfont=normalfont,labelsep=colon,strut=off} 
\floatstyle{ruled}
\newfloat{listing}{tb}{lst}{}
\floatname{listing}{Listing}
\title{Don't Start Over: A Cost-Effective Framework for Migrating Personalized Prompts Between LLMs}

\author{
    Ziyi Zhao\textsuperscript{\rm 1}, 
    Chongming Gao\textsuperscript{\rm 1}\thanks{Corresponding authors.},
    Yang Zhang\textsuperscript{\rm 2},
    Haoyan Liu\textsuperscript{\rm 1}\footnotemark[1],
    Weinan Gan\textsuperscript{\rm 3},
    Huifeng Guo\textsuperscript{\rm 3},
    Yong Liu\textsuperscript{\rm 3},
    Fuli Feng\textsuperscript{\rm 1},
}
\affiliations{
    \textsuperscript{\rm 1}University of Science and Technology of China, Hefei, China\\
    \textsuperscript{\rm 2}National University of Singapore, Singapore\\
    \textsuperscript{\rm 3}Huawei Technologies Co., Ltd, Shenzhen, China\\
    re2477036@mail.ustc.edu.cn, {chongming.gao,zyang1580}@gmail.com, liuhaoyan@ustc.edu.cn, {ganweinan1,huifeng.guo,liu.yong6}@huawei.com, fulifeng93@gmail.com\\
}

\begin{document}

\maketitle

\begin{abstract}
Personalization in Large Language Models (LLMs) often relies on user-specific soft prompts. However, these prompts become obsolete when the foundation model is upgraded, necessitating costly, full-scale retraining. To overcome this limitation, we propose the \textbf{P}rompt-level \textbf{U}ser \textbf{M}igration \textbf{A}dapter (\textbf{PUMA}), a lightweight framework to efficiently migrate personalized prompts across incompatible models. PUMA utilizes a parameter-efficient adapter to bridge the semantic gap, combined with a group-based user selection strategy to significantly reduce training costs. Experiments on three large-scale datasets show our method matches or even surpasses the performance of retraining from scratch, reducing computational cost by up to 98\%. The framework demonstrates strong generalization across diverse model architectures and robustness in advanced scenarios like chained and aggregated migrations, offering a practical path for the sustainable evolution of personalized AI by decoupling user assets from the underlying models. 

\end{abstract}

\begin{links}
\link{Code}{https://github.com/Kimagure7/Dont-Start-Over}
\end{links}

\section{Introduction}

The rapid advancement of large language models (LLMs) \cite{vaswani2023attentionneed, brown2020languagemodelsfewshotlearners, chowdhery2022palmscalinglanguagemodeling,openai2024gpt4technicalreport} has fundamentally reshaped the landscape of NLP, demonstrating remarkable capabilities across a multitude of tasks. As these powerful models are increasingly integrated into real-world applications, the focus is shifting from general-purpose utility to deep personalization -- a critical step to meet user expectations for experiences tailored to their unique needs and preferences \cite{zhang2025personalizationlargelanguagemodels, Chen_2024, salemi2024lamplargelanguagemodels}. This evolution is prominent in burgeoning domains such as personal assistants \cite{10.1145/3580305.3599572,zhang-etal-2024-llm-based}, adaptive education \cite{wang2024largelanguagemodelseducation}, and recommendation systems \cite{gao2024sprec,fan2025fine,gao2025flower,cai-etal-2025-k,zhang2024collmintegratingcollaborativeembeddings,wentao_sigir}.

Soft prompts \cite{lester2021powerscaleparameterefficientprompt} have emerged as a key technology for realizing this deep personalization. The core idea is to encode and carry each user's unique preferences and knowledge within a lightweight, efficient, and non-invasive dedicated vector -- the \emph{soft prompt}. This approach has given rise to a new application architecture: a ``1+N'' hybrid system \cite{tan2025democratizinglargelanguagemodels, li2023personalizedpromptlearningexplainable}. Such a system combines a single, powerful, general-purpose foundation model (the ``1'') with thousands or even millions of independent soft prompts (the ``N''), where each prompt represents an individual user's personalized representation.
This method allows for efficient, user-specific customization without altering the parameters of the LLM model.


\begin{figure}[tbp]
    \centering
    \includegraphics[width=\columnwidth]{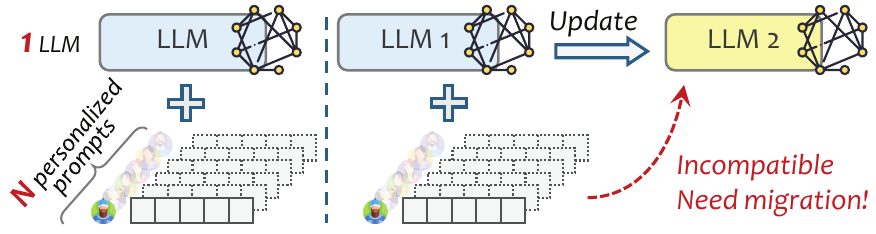}
    \caption{The ``1+N'' system (left) and an illustration of the migration of personalized prompts to a new LLM (right).}
    \label{fig:opening}
\end{figure}


However, this approach introduces a critical vulnerability: the efficacy of soft prompts is inherently tied to the specific foundation model they were trained on. The lifecycle of these ``1+N'' systems inevitably involves replacing the foundation model, whether upgrading to a more powerful successor or adopting a smaller, more efficient variant to meet new deployment constraints. Any such replacement shatters the semantic alignment between the prompts and the model, rendering the entire corpus of personalized soft prompts obsolete.  Consequently, the valuable personalization accumulated across the user base is invalidated, necessitating a complete and cost-prohibitive retraining of all prompts from scratch. This challenge motivates the central question of our work: 
\emph{Can we migrate a large corpus of personalized soft prompts from a source model to a new target model with high fidelity, but at a fraction of the computational cost of full retraining?}

Migrating user-level personalization presents a distinct challenge compared to existing soft prompt transfer paradigms. Most research has focused on the task level, where the goal is to transfer a single, public prompt trained on one task (e.g., natural language inference) to accelerate fine-tuning on a different target task (e.g., text classification) \cite{vu2022spotbetterfrozenmodel, asai2022attemptparameterefficientmultitasktuning}. In this case, the knowledge is impersonal and designed for cross-task use. In contrast, our work focuses on migrating thousands of private, user-specific soft prompts, each tailored to an individual.


To tackle the challenge of personalized soft prompt migration, we decompose it into two coupled subproblems: \emph{semantic incompatibility} -- how to enable the target model to interpret prompts trained on a different source model -- and \emph{migration efficiency} -- how to scale migration to support tens of thousands of users with minimal cost.

We propose \textbf{P}rompt-level \textbf{U}ser \textbf{M}igration \textbf{A}dapter (\textbf{PUMA}), a lightweight framework designed for both semantic alignment and scalable migration. It comprises two key components: (1) a parameter-efficient adapter trained end-to-end to bridge semantic gaps between models; and (2) a user selection strategy that groups users by their prompt embeddings and output variance to form a small, representative training subset, significantly reducing cost without sacrificing performance.

We evaluate our framework on the task of personalized recommendation, focusing on migration across models of varying sizes (e.g., LLaMA 1B $\rightarrow$ 3B) and even across model families (e.g., LLaMA $\rightarrow$ Qwen). Experiments on three large-scale datasets demonstrate that our method matches or surpasses the performance of costly from-scratch retraining, while reducing computation by up to 98\%. These results highlight the scalability and efficiency of our approach for real-world personalization.

We further extend our method to two advanced migration paradigms: \emph{chained migration}, where a soft prompt is sequentially migrated across multiple target LLMs, and \emph{aggregated migration}, where soft prompts from multiple base models are fused into a single target model. The results show that leveraging multiple sources, especially in the aggregated setting, enhances knowledge integration and leads to improved post-migration performance, offering valuable insights for scalable user adaptation.

In summary, our main contributions are threefold:

\begin{itemize}
    \item We are the first to identify and formalize the challenge of migrating user-level personalized soft prompts across foundation models.
    
    \item We propose \textit{PUMA}, a novel, lightweight framework that uses an end-to-end trained adapter for effective migration, along with a group-based user sampling strategy to significantly enhance efficiency.
    
    \item Experiments on three large-scale datasets show our approach matches and often surpasses the performance of retraining from scratch, but with substantially lower computational cost.
\end{itemize}

\section{Related Work}

\subsection{Personalization in Large Language Models}

The demand for LLMs to cater to individual user needs and preferences has grown significantly \cite{liu2025surveypersonalizedlargelanguage}. To meet this demand, researchers have explored various techniques, such as retrieval-augmented generation \cite{gao2024retrievalaugmentedgenerationlargelanguage,salemi2024lamplargelanguagemodels}, prompt engineering \cite{kang2023llmsunderstanduserpreferences, liu2023chatgptgoodrecommenderpreliminary}, and reinforcement learning \cite{jang2023personalizedsoupspersonalizedlarge}. Among these, parameter-efficient fine-tuning (PEFT) methods have become a prominent approach for personalization \cite{lester2021powerscaleparameterefficientprompt, hu2021loralowrankadaptationlarge}. \citet{tan2025democratizinglargelanguagemodels} employed personalized PEFT modules to capture user preferences. \citet{huang2024selectivepromptingtuningpersonalized} proposed selective prompt tuning to achieve a personalized dialogue by adaptively selecting the appropriate soft prompts.

While effective, these PEFT personalization methods are coupled to their base models, requiring costly retraining upon model updates.  To the best of our knowledge, our work is the first to investigate the migration of these user-level personalized parameters across different LLMs.

\subsection{Coreset Selection}

Our user selection strategy is an approach to coreset selection, a field whose primary objective is to identify a small representative data subset (a ``coreset'') to approximate training on a full dataset \cite{mirzasoleiman2020coresetsdataefficienttrainingmachine}.  Existing methodologies largely fall into two categories: score-based and optimization-driven \cite{albalak2024surveydataselectionlanguage}. Score-based methods rank data points using various metrics. For instance, some approaches use geometric properties like k-means clustering to select a diverse set \cite{sorscher2023neuralscalinglawsbeating}, while others prioritize ``informative'' examples identified by high prediction uncertainty or training loss \cite{coleman2020selectionproxyefficientdata, zheng2023coveragecentriccoresetselectionhigh}. Optimization-driven methods, in contrast, select a coreset by matching its training gradients to those of the full dataset, using first-order \cite{mirzasoleiman2020coresetsdataefficienttrainingmachine} or more refined second-order \cite{pooladzandi2022adaptivesecondordercoresets} information.


However, approaches requiring a per-sample evaluation pass, such as those based on gradients or uncertainty, are computationally infeasible at our scale. We therefore propose an efficient selection strategy that circumvents this specific bottleneck.

\subsection{Knowledge Transfer in Soft Prompt}
Research on the transferability of soft prompts has primarily focused on improving training efficiency and model generalization across different downstream tasks. For instance, \citet{Su_2022} empirically studied the transferability of soft prompts across different downstream tasks and pre-trained language models (PLMs). \citet{vu2022spotbetterfrozenmodel} utilized a retrieval-based strategy that measures task similarity to select the most relevant prompt as the initialization for a new task. Similarly, \citet{asai2022attemptparameterefficientmultitasktuning} introduced an attention-based mechanism to mix multiple prompts for a new task.

Our work, however, addresses a different problem. These prior studies focus on transferring general, task-level knowledge --- a ``one-to-one'' or ``few-to-one'' transfer problem. In contrast, we tackle the ``N-to-N'' challenge of migrating a large corpus of user-personalized prompts, preserving them as valuable assets across foundation model changes.

\section{Methodology}\label{sec:methodology}

We will first formalize the cross-model migration problem, then present our Adapter-based migration framework, and introduce the group-based user selection strategy designed to make this process efficient.

\subsection{Problem Formulation}\label{subsec:problem_formulation}

To clearly delineate the scope of our research, we first formalize the core concepts and objectives.

\subsubsection{Personalization with Soft Prompts}


\begin{figure}[tbp]
    \centering
    \includegraphics[width=\columnwidth]{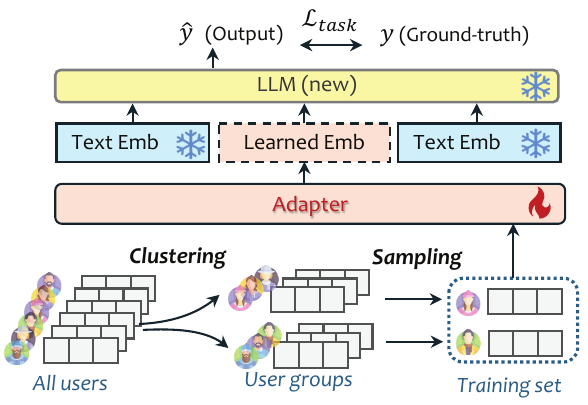}
    \caption{Illustration of PUMA, consisting of a group-based user selection strategy and a migration adapter. Users are first clustered via $K$-means on personalized embeddings, then sub-grouped by output variance, from which training users are sampled.}
    \label{fig:framework} 
\end{figure}



Consider a frozen, pre-trained source foundation model, $M_s$, and a set of users $\mathcal{U} = \{u_1, u_2, \dots, u_{|\mathcal{U}|}\}$ and items $\mathcal{I} = \{i_1, i_2, \dots, i_{|\mathcal{I}|}\}$. To capture individual user preferences, we learn a unique soft prompt, $p_u \in \mathbb{R}^{l \times d_s}$, for each user $u \in \mathcal{U}$. Here, $l$ is the prompt length and $d_s$ is the embedding dimensionality of the source model $M_s$.

The goal is to optimize the entire collection of user prompts, $\{p_u\}_{u \in \mathcal{U}}$, by minimizing a task-specific loss function. This process trains the prompts to elicit personalized predictions from the frozen model $M_s$. The optimization is formally expressed as:
\begin{equation}\label{eq:source_optimization}
\argmin_{ \{p_u\}_{u \in U} } \sum_{(u, i, y) \in \mathcal{D}} \mathcal{L}_{\text{task}}\left( M_s\left( T(p_u, \phi(i)) \right), y\right),
\end{equation}
where $\mathcal{D}$ is the training dataset containing user-item-outcome tuples $(u, i, y)$, $\phi(i)$ is the textual representation of an item, and $T$ is a template that structures the user prompt $p_u$ and item text for the model. $\mathcal{L}_{\text{task}}$ measures the discrepancy between the model's prediction and the ground truth $y$ (e.g., a rating).

\subsubsection{Cross-Model Prompt Migration}


The primary challenge occurs when the source model $M_s$ is replaced with a target model $M_t$, especially when $M_t$ has a different architecture or embedding dimension ($d_t \neq d_s$), where $d_t$ is the embedding dimensionality of the target model. This mismatch creates a semantic gap, making the original prompts $\{p_u\}$ incompatible.


To address this, our goal is to learn a lightweight migration function, $\Phi$, parameterized by $\theta$. This function maps each source prompt $p_u$ to a functionally equivalent prompt $p'_u$ for the target model:
\begin{equation*}
    p'_u = \Phi_\theta(p_u), \quad \text{where } p_u \in \mathbf{R}^{l \times d_s} \text{ and } p'_u \in \mathbf{R}^{l \times d_t}
\end{equation*}
The optimal parameters $\theta^*$ are learned by minimizing the task loss on the target model. During this optimization, both the target model $M_t$ and the source prompts $\{p_u\}$ remain frozen, so only the migration function $\Phi$ is trained:
\begin{equation}\label{eq:migration_optimization_split}
\begin{array}{l@{}l}
    \theta^* = \displaystyle\argmin_{\theta}  \displaystyle\sum_{(u, i, y) \in D} 
     \mathcal{L}_{\text{task}}\left( M_t\left( T(\Phi_\theta(p_u), \phi(i)) \right), y \right)
\end{array}
\end{equation}

\subsection{Adapter Framework}\label{subsec:adapter_framework}

We implement the migration function $\Phi$ as a lightweight adapter -- a feed-forward network with residual connections and Layer Normalization. This architecture balances the representational capacity with low computational cost.

Crucially, this adapter is trained end-to-end by optimizing the task-specific loss in Eq.~\eqref{eq:migration_optimization_split}, ensuring that the learned transformation is functional and preserves the downstream utility of each personalized prompt.


\subsubsection{Extension to Advanced Migration Topologies.} The flexibility of PUMA's design naturally extends to more complex migration scenarios that mirror real-world system evolution. We investigate two such advanced topologies: \emph{chained migration} and \emph{aggregated migration}. Chained migration ($M_A \rightarrow M_B \rightarrow M_C$) handles sequential model changes. Aggregated migration fuses personalization from multiple source models into a single target ($[M_A, M_B] \rightarrow M_C$), a common need after system mergers or A/B testing. We achieve this by concatenating a user's source prompts (e.g., $[p_u^A; p_u^B]$) and mapping the composite vector to the target model, thereby synthesizing a richer user representation from diverse sources. We empirically validate both topologies in our experiments.

\subsection{Efficient Migration via Group-based User Selection}


To make the migration computationally tractable for large user populations,  we train the adapter on a small data subset $\mathcal{D}' = \{ (u, i, y) \in \mathcal{D} \mid u \in \mathcal{U}'\}$. The core challenge lies in constructing a compact user subset $\mathcal{U}'$ that is highly representative of the entire user population.


Our selection strategy pivots on a key insight: an ideal user subset must embody both the diversity of user preferences and the spectrum of their complexity. To capture preference diversity, we leverage the geometric structure of the source prompt embeddings. For complexity, we posit that a user's historical output variance serves as an effective and computationally cheap proxy. Users with low variance exhibit consistent, easily modeled preferences (e.g., always giving high ratings). In contrast, high-variance users display more nuanced tastes, posing a greater learning challenge. A robust migration function must be trained on examples spanning this entire spectrum of complexity.
We implement this strategy through a two-stage selection process:

    \paragraph{Stage 1: Cluster for diversity.} 
    We apply $K$-means clustering to the source prompt $\{p_u\}_{u \in \mathcal{U}}$. This partitions users into $k$ distinct clusters, ensuring that our selection captures a wide array of learned preference profiles.
    
    \paragraph{Stage 2: Grouped sampling by variance.}
    Within each cluster, we stratify users by output variance and then sample from these bins, using a normal distribution to weight the medium-variance groups more heavily.

\section{Experiments}

Our experiments answer four research questions (RQs):
\textbf{RQ1}: How effective is PUMA at migrating user personalization compared to the costly baseline of full retraining?
\textbf{RQ2}: How efficiently does our group-based user selection strategy reduce computational costs while preserving performance, relative to random sampling?
\textbf{RQ3}: Does PUMA generalize across diverse model architectures and families?
\textbf{RQ4}: How robust is PUMA when applied to advanced migration topologies, such as chained and aggregated migration?

\subsection{Experimental Setup}
    
    
\subsubsection{Datasets.}
We utilize three large-scale datasets (see Table~\ref{tab:dataset_stats} for statistics).
\textbf{Amazon (Movies \& TV)} \cite{hou2024bridging}: A subset of the Amazon Product Review dataset\footnote{https://amazon-reviews-2023.github.io/} containing 1-to-5 star ratings, used for explicit rating prediction.
\textbf{Yelp}\footnote{https://business.yelp.com/data/resources/open-dataset/}: A dataset of 1-to-5 star ratings for local businesses (e.g., restaurants), modeling real-world consumer preferences.
\textbf{MIND} \cite{wu-etal-2020-mind}: A news recommendation dataset based on implicit feedback (clicks) to predict click-through rate (CTR).

\begin{table}[ht]
\centering
\begin{tabular}{lccc}
\toprule
\textbf{Statistics} & \textbf{Amazon} & \textbf{MIND} & \textbf{Yelp} \\
\midrule
\#Users & 30,287 & 50,000 & 32,850 \\
\#Items & 96,636 & 19,368 & 129,076 \\
\#Records & 1,166,752 & 3,892,068 & 1,543,687 \\
Records/User & 38.52 & 77.84 & 46.99 \\
Sparsity & 99.96\% & 99.60\% & 99.96\% \\
Positive Ratio & N/A & 17.80\% & N/A \\
\bottomrule
\end{tabular}
\caption{Statistics of the datasets used in our experiments.}
\label{tab:dataset_stats}
\end{table}

\subsubsection{Evaluation Metrics.}

We evaluate performance using standard metrics appropriate for each task. For the rating prediction tasks on Amazon and Yelp, we use root mean square error (RMSE) and mean absolute error (MAE). For the click-through rate (CTR) prediction task on MIND, we report the area under the ROC curve (AUC) and user-weighted AUC (uAUC), which is the average of AUC scores computed for each individual user.


\subsubsection{Implementation Details.}

To address RQ1 and RQ2, we simulate a common model upgrade scenario, migrating personalized prompts from a Llama-2-1B-Instruct source model to a more powerful Llama-2-3B-Instruct target \cite{grattafiori2024llama3herdmodels}. To assess generalization and robustness (RQ3 and RQ4), we conduct migrations across a diverse portfolio of five models: Llama3.2-3B-Instruct, Qwen2.5-3B-Instruct \cite{qwen2,qwen2.5}, Stablelm-2-1\_6b-chat \cite{StableLM-2-1.6B}, Phi-3-mini-4k-instruct \cite{abdin2024phi3technicalreporthighly}, and Gemma-3-1b-it \cite{gemma_2025}.

We conducted all experiments on NVIDIA A100 GPUs using PyTorch 2.5. We leverage foundation models from the Hugging Face Hub, keeping their parameters frozen throughout all training phases. We first pre-train the user-specific soft prompts (length $l=1$) on the source model for 15 epochs with a learning rate of $5 \times 10^{-4}$. Subsequently, the PUMA adapter is trained for 4 epochs using the FusedAdam optimizer with a learning rate set to $1 \times 10^{-4}$ and a batch size set to 32.


We tailor the loss function to the specific task of each dataset. For the rating prediction tasks (Amazon and Yelp), we employ a hybrid loss objective. We extract the logits corresponding to the five discrete rating tokens (``1'' through ``5'') from the LLM's output. A cross-entropy loss ($\mathcal{L}_{CE}$) is applied to these logits to treat rating prediction as a classification problem. Concurrently, we feed the same logits into a lightweight MLP head to regress a continuous rating value, which is supervised by a mean squared error loss ($\mathcal{L}_{MSE}$). The final objective is a weighted sum: $0.8 \cdot \mathcal{L}_{MSE} + 0.2 \cdot \mathcal{L}_{CE}$ \cite{10.1145/3716499}. For the binary CTR prediction task on the MIND dataset, we employ a standard binary cross-entropy loss ($\mathcal{L}_{BCE}$) computed on the logit of the ``yes'' token \cite{zhang2024collmintegratingcollaborativeembeddings}.

\begin{table*}[ht]
\centering
\begin{tabular*}{\textwidth}{@{\extracolsep{\fill}}lcrrrrrr@{}}
\toprule
\textbf{Method} & \textbf{Trainable} & \multicolumn{2}{c}{\textbf{Amazon}} & \multicolumn{2}{c}{\textbf{MIND}} & \multicolumn{2}{c}{\textbf{Yelp}} \\ 
\cline{3-4} \cline{5-6} \cline{7-8}
& \textbf{Params (M)} & RMSE$\downarrow$ & MAE$\downarrow$ & AUC$\uparrow$ & uAUC$\uparrow$ & RMSE$\downarrow$ & MAE$\downarrow$ \\ 
\midrule
\multicolumn{8}{l}{\textit{Baselines}} \\
Full Retraining         & N/A   & 0.9414          & 0.6296          & 0.5778          & 0.5289          & 1.1994          & 0.9269          \\
Source Model Performance& N/A   & 0.9438          & 0.6306          & 0.5742          & 0.5312          & 1.2005          & 0.9369          \\
Random Initialization   & N/A   & 1.2352          & 1.1168          & 0.4917          & 0.4883          & 1.6671          & 1.4981          \\ 
\midrule
\textbf{PUMA}           & 88.1  & \textbf{0.9135} & \textbf{0.5701} & \textbf{0.6546} & \textbf{0.6552} & \textbf{1.1073} & \textbf{0.8493} \\
\bottomrule
\end{tabular*}
\caption{Performance comparison of different methods for prompt migration. The best results for each metric are highlighted in bold. The arrows $\downarrow$ indicate that lower is better, while $\uparrow$ indicates that higher is better. PUMA was trained using data from the entire user population in this experiment.}
\label{tab:adapter_architectures}
\end{table*}

\subsubsection{Compared Methods.}
We benchmark PUMA against two fundamental baselines to establish performance boundaries (RQ1). \textbf{Full Retraining} serves as the performance upper bound, where all user prompts are retrained from scratch on the target model ($M_t$). Conversely, \textbf{Random Initialization} provides a performance lower bound by using randomly initialized vectors. To ensure a fair comparison in rating prediction tasks, the task-specific MLP head is still trained.

For evaluating efficiency (RQ2), we compare our group-based selection strategy against a comprehensive suite of alternative sampling methods:
\begin{itemize}
    \item \textbf{Simple Baselines:} We start with Random Sampling, which selects users uniformly at random. We also test simple single-heuristic methods like Variance Bucketing and Loss Bucketing, which stratify the user pool by either historical output variance or task loss on the source model before sampling.
    \item \textbf{Clustering-based Strategies:} These methods leverage the geometry of the prompt embeddings. $K$-means Stratified clusters users and samples proportionally from each cluster. A variant, $K$-means with PCA, first applies a dimensionality reduction step to the prompt embeddings prior to clustering.
    \item \textbf{Hybrid and Feature-based Clustering Strategies:} We test hybrid methods like $K$-means + FPS, which applies Farthest Point Sampling within clusters for diversity, and K-Means + Loss Stratification, which subgroups clusters by source model loss. We also evaluate $K$-means (on FFN Activ.), a strategy that clusters users based on richer feature vectors instead of on soft prompts; these are extracted by feeding each soft prompt into the source model and concatenating the activations from its final three FFN layers \cite{Su_2022}.
\end{itemize}

\subsection{Performance Results}

We now present the empirical evaluation of PUMA, systematically addressing its effectiveness, efficiency, generalization, and robustness.

\subsubsection{Effectiveness (RQ1)}

The results presented in Table~\ref{tab:adapter_architectures} clearly establish PUMA's effectiveness. Across all three datasets, our framework not only successfully migrates user personalization but consistently outperforms the strong baseline of a full, from-scratch retraining. On the rating prediction tasks, PUMA achieves a lower RMSE on both Amazon (0.9135 vs. 0.9414) and Yelp (1.1073 vs. 1.1994). This superior performance is also evident on the MIND dataset, where PUMA significantly improves the uAUC from 0.5289 (full retraining) to 0.6552, demonstrating its efficacy across diverse task formulations.

\begin{table}[t]
    \centering
    \resizebox{\linewidth}{!}{
        \setlength{\tabcolsep}{4pt} 
        \begin{tabular}{lrrrr}
        \toprule
        \textbf{Method} & \textbf{h/Epoch} & \textbf{Epochs} & \textbf{Total (h)} & \textbf{Speedup} \\
        \midrule
        Full Retraining & 3.00 & 8 & 24.0 & 1x \\
        PUMA (2k users) & 0.16 & 3 & \textbf{0.48} & \textbf{50x} \\
        \bottomrule
        \end{tabular}
    }
    \caption{Efficiency comparison on the Amazon dataset.}
    \label{tab:efficiency_headline}
\end{table}

Notably, PUMA's consistent outperformance suggests a fundamental advantage over retraining from scratch. We hypothesize that this stems from reframing the problem: instead of learning thousands of individual user representations in isolation, PUMA learns a single, supervised mapping function. This shared adapter appears to discover a more generalized and robust transformation into the target model's semantic space. The power of this generalized mapping is further explored in our analysis of advanced migration scenarios (RQ4).

\begin{figure}[!tbp]
    \centering 
    \includegraphics[width=\columnwidth]{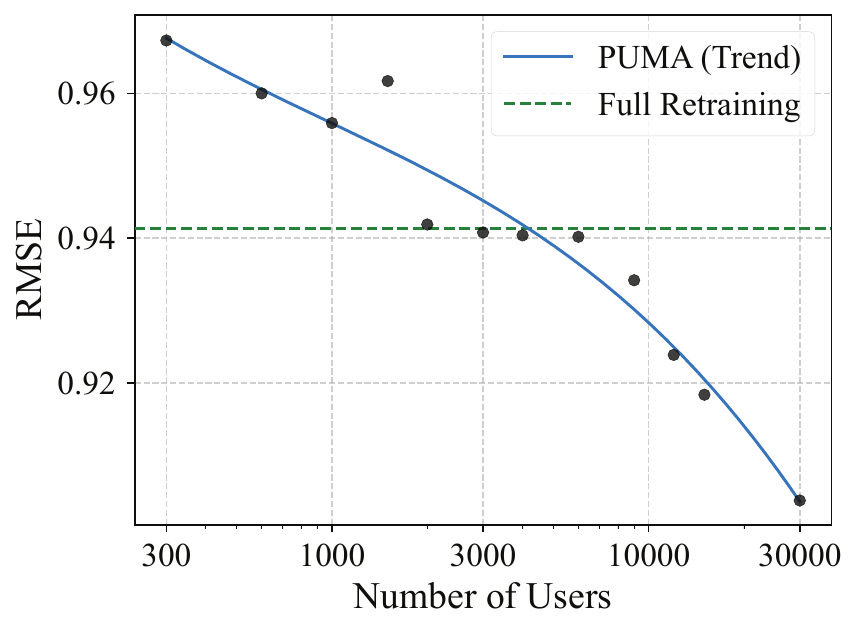}
    \caption{Performance of random user sampling on Amazon.}
    \label{fig:user_scale_trend} 
\end{figure}



\begin{table*}[t]
\centering
\begin{tabular*}{\textwidth}{@{\extracolsep{\fill}}lcccccc}
\toprule
\textbf{Method} & \multicolumn{2}{c}{\textbf{Amazon}} & \multicolumn{2}{c}{\textbf{MIND}} & \multicolumn{2}{c}{\textbf{Yelp}} \\
\cline{2-3} \cline{4-5} \cline{6-7}
~ & \textbf{RMSE $\downarrow$} & \textbf{MAE $\downarrow$} & \textbf{AUC $\uparrow$} & \textbf{uAUC $\uparrow$} & \textbf{RMSE $\downarrow$} & \textbf{MAE $\downarrow$} \\
\midrule
\multicolumn{7}{l}{\textit{Baselines}} \\
Random & 0.9419 & 0.6155 & 0.5916 & 0.5861 & 1.1146 & 0.8627 \\
Random (6k) & 0.9320 & 0.6098 & 0.6585 & 0.6636 & 1.1128 & 0.8578 \\
\midrule
\multicolumn{7}{l}{\textit{Single-Heuristic Selection}} \\
Variance Bucketing & 0.9508 & 0.6350 & 0.5934 & 0.5888 & 1.1171 & 0.8666 \\
Loss Bucketing & 0.9454 & 0.6209 & 0.5623 & 0.5845 & 1.1127 & \textbf{0.8595} \\
K-Means Stratified & 0.9546 & \textbf{0.5885} & 0.5830 & 0.5927 & 1.1152 & 0.8618 \\
K-Means with PCA & \underline{0.9351} & 0.6005 & 0.5608 & 0.5652 & 1.1138 & 0.8659 \\
\midrule
\multicolumn{7}{l}{\textit{Hybrid Selection Strategies (on Prompt Embeddings)}} \\
K-Means + FPS & 0.9355 & 0.6113 & 0.5990 & 0.5966 & 1.1147 & 0.8735 \\
K-Means + Loss Stratification & 0.9371 & 0.6065 & 0.5779 & 0.5884 & 1.1122 & 0.8634 \\
\textbf{K-Means + Variance Stratification (PUMA)} & \textbf{0.9315} & \underline{0.5986} & \textbf{0.6346} & \textbf{0.6344} & \textbf{1.1111} & \underline{0.8616} \\
\midrule
\multicolumn{7}{l}{\textit{Ablation: Strategies using FFN Activations}} \\
K-Means (on FFN Activ.) & 0.9373 & 0.6096 & 0.5624 & 0.5697 & \underline{1.1112} & 0.8690 \\
K-Means (on FFN Activ.) + Loss Stratification & 0.9467 & 0.6339 & \underline{0.6152} & \underline{0.6298} & 1.1115 & 0.8633 \\
K-Means (on FFN Activ.) + Variance Stratification & 0.9365 & 0.6043 & 0.5877 & 0.5936 & 1.1184 & \underline{0.8616} \\
\bottomrule
\end{tabular*}
\caption{Performance comparison of user selection strategies. The best result is in \textbf{bold} and the second-best is \underline{underlined}. Baselines are excluded from highlighting. $\downarrow$ indicates lower is better; $\uparrow$ indicates higher is better. \textit{Note}: Except for Random (6K), which was trained on 6,000 users, all strategies were trained under a fixed budget of 2,000 users for Amazon/Yelp and 1,500 users for MIND.}
\label{tab:user_selection_results_final_v4}
\end{table*}

\subsubsection{Efficiency (RQ2)}



Our evaluation of user selection strategies reveals the remarkable efficiency of the PUMA framework. As shown in Table~\ref{tab:user_selection_results_final_v4}, our ``K-Means + Variance Stratification (PUMA)'' approach is highly cost-effective. By operating under a strict computational budget (2,000 users for Amazon/Yelp, 1,500 for MIND), it not only surpasses random sampling at the same scale but consistently matches or exceeds the performance of baselines that use 3-4 times more data. For example, on the Amazon dataset, our method with 2,000 users (RMSE 0.9315) outperformed the ``Random (6k)'' baseline (RMSE 0.9320), despite using only one-third of the users. This efficiency is even more pronounced when compared to full retraining, where PUMA achieves a remarkable 50x speedup, as detailed in Table~\ref{tab:efficiency_headline}.

Figure~\ref{fig:user_scale_trend} starkly illustrates the inefficiency of naive random sampling. The performance trend reveals that approximately 5,000 randomly selected users are required merely to match the RMSE of a full, from-scratch retraining. This substantial data requirement highlights the prohibitive computational cost of an unguided sampling approach and firmly establishes the need for our more sophisticated, group-based selection strategy to achieve high-fidelity migration efficiently.



\subsubsection{Generalization across Architectures (RQ3)}
To evaluate PUMA's generalization, we migrated prompts across diverse model architectures and families. For these and the subsequent experiments in RQ4, we kept the training set fixed at 6,000 users. To uniformly assess the impact of model migration across different model families, we define the relative improvement, or gain, as the ratio of the RMSE of the fully retrained model to the RMSE of the migrated model.

\begin{figure}[!ht]
    \centering 
    \includegraphics[width=0.99\columnwidth]{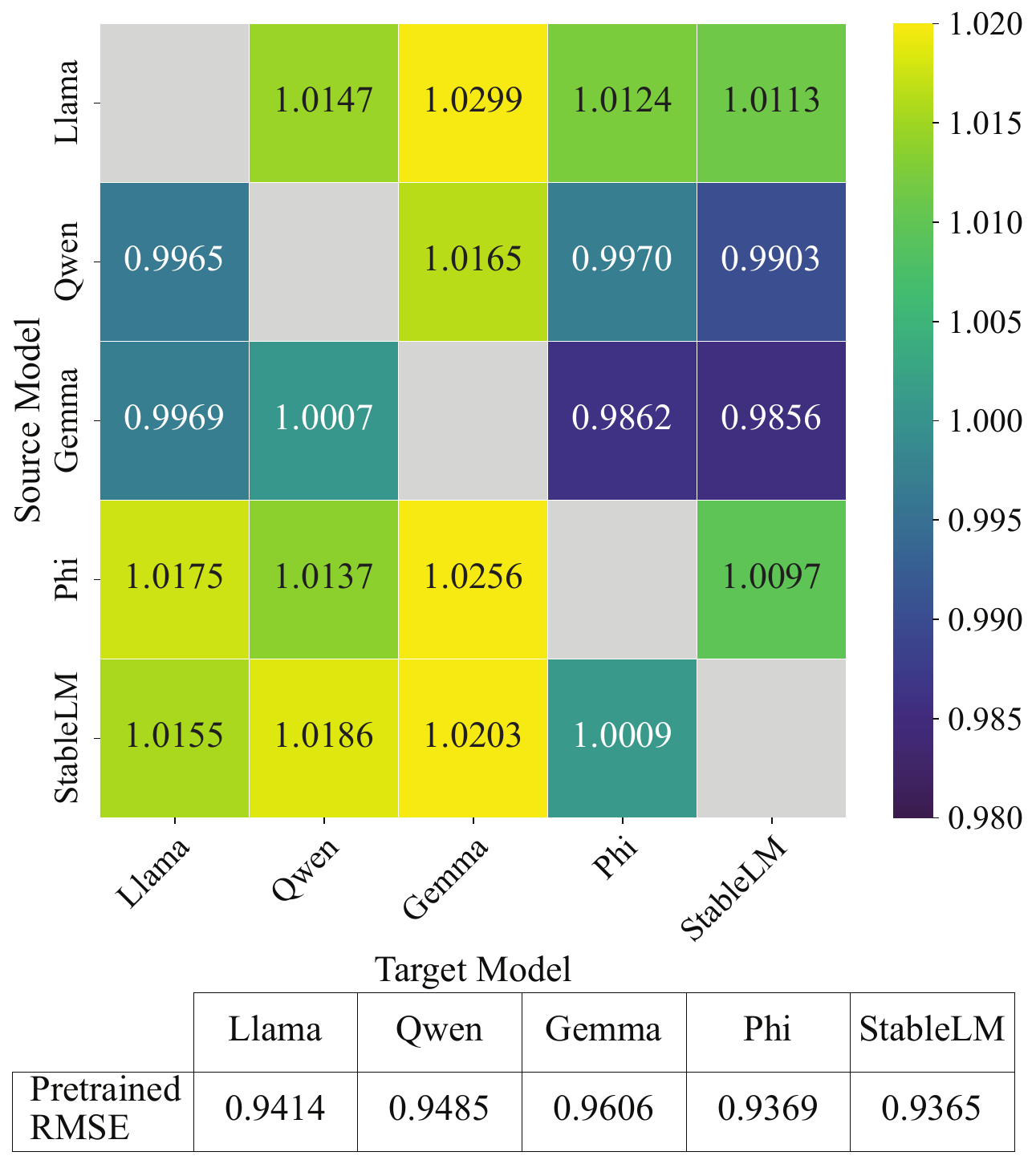}
    \caption{Performance gain heatmap for different architectures. (Gain = full retrained RMSE / migrated RMSE)}
    \label{fig:perf_matrix} 
\end{figure}

The heatmap in Figure~\ref{fig:perf_matrix} confirms that PUMA generalizes remarkably well. Migrations between distinct model families consistently yield substantial performance gains, often matching or even exceeding those of a full retraining. As expected, migration efficacy is bounded by the quality of the source prompts. For instance, migrating from a weaker source model (e.g., Gemma-3) to a stronger target yields significant improvement, though it may not reach the performance of retraining from scratch on that target. Conversely, migrating from a high-performing source (e.g., Phi-3) can surpass it.

This outcome is logical, as the migration process transfers existing knowledge but does not create it. This observation motivates our final question: can the PUMA framework fuse knowledge from multiple sources to achieve even richer personalization? We explore this possibility in RQ4.

\subsubsection{Advanced Migration Scenarios (RQ4)}
In this section, we explore two advanced migration scenarios: chain migration and aggregated migration.
\paragraph{(1) Robustness in Chain Migration.} 
To test our framework's robustness against cumulative error from successive model changes, we conducted a chain migration across five models: Llama3.2 $\rightarrow$ Qwen2.5 $\rightarrow$ Gemma-3 $\rightarrow$ StableLM-2 $\rightarrow$ Phi-3. At each step, the newly migrated prompts become the source for the next migration, without any retraining on the original user data.

\begin{figure}[tbp]
    \centering 
    \includegraphics[width=\columnwidth]{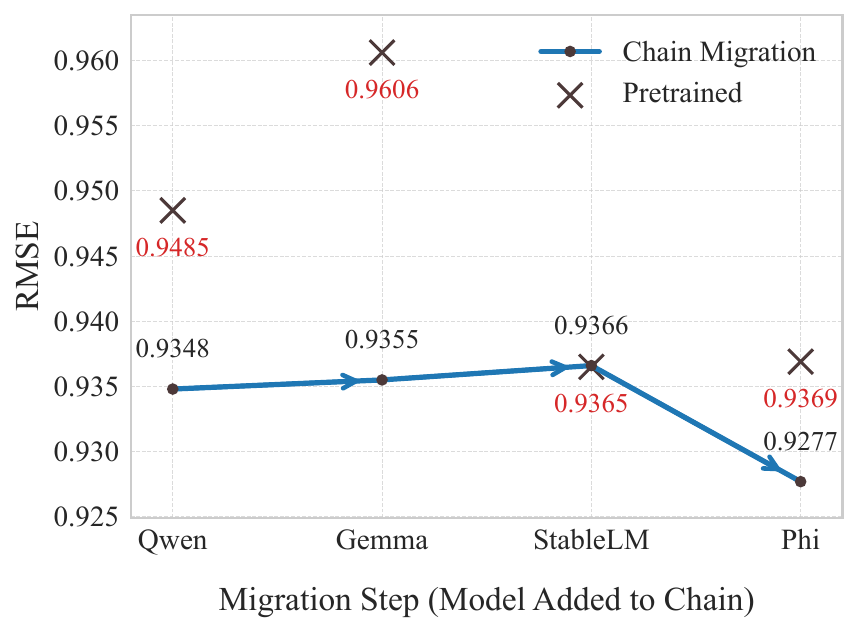}
    \caption{Stability in chain migration (started from Llama).}
    \label{fig:chain_migration} 
\end{figure}

The results, shown in Figure~\ref{fig:chain_migration}, demonstrate remarkable stability. The performance (RMSE) remains consistently strong throughout the entire chain, starting at 0.9348 and ending at 0.9277. This result not only outperforms retraining from scratch at each individual stage but also validates PUMA's robust ability to preserve user personalization across multiple successive model updates.

We do observe a minor performance dip when migrating from the Gemma model. This is expected, as some information loss is likely when transitioning to a comparatively weaker model in the chain. Nevertheless, the subsequent migration step still achieves a final performance level superior to that of a direct migration from a pretrained Gemma to StableLM (as shown in Figure~\ref{fig:perf_matrix}), underscoring the framework's robustness.

\paragraph{(2) Knowledge Fusion from Multiple Sources.} 
We then investigated PUMA's capacity for knowledge fusion through an aggregated migration scenario. In this experiment, we combined prompts from two distinct source models and migrated them to a single target model, which we set as Phi-3.

\begin{figure}[tbp]
    \centering 
    \includegraphics[width=\columnwidth]{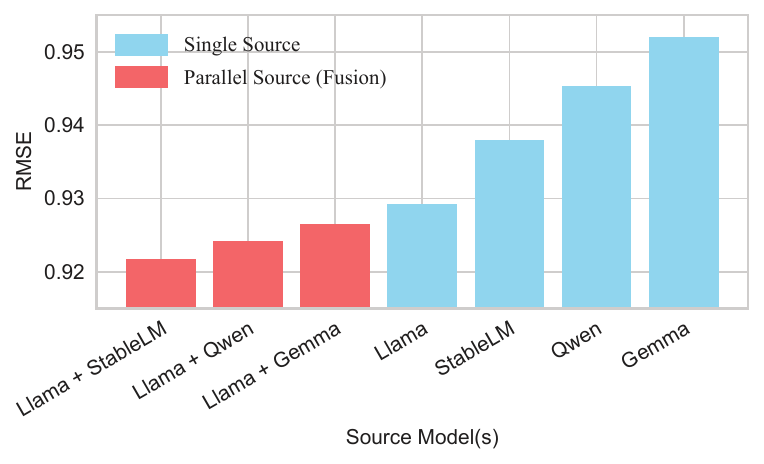}
    \caption{Performance of aggregated migration from one or two source models to Phi-3.}
    \label{fig:aggregated_migration}
\end{figure}

As illustrated in Figure~\ref{fig:aggregated_migration}, the aggregated migration strategy consistently outperforms migrations from any single source model. For instance, by fusing prompts from ``Llama + StableLM'', the resulting model achieves an RMSE of 0.9217. This marks a substantial improvement over migrating solely from Llama (0.9293) or StableLM (0.9380). The combination of these two models yielded the best overall performance, likely because they were also the strongest individual performers in prior migration tests.


This consistent improvement highlights a principle of knowledge synergy: different foundation models capture complementary aspects of user preferences. By providing the adapter with a composite representation drawn from these diverse sources, the migration process becomes more effective. This finding reframes prompt migration from a maintenance task into a strategic opportunity to enhance personalization. It suggests a cumulative approach, where personalized assets are continuously enriched by integrating knowledge from multiple model ecosystems, evolving into dynamic, improvable profiles rather than static parameters tied to a single platform.

\section{Conclusion \& Discussion}

In this paper, we address the critical challenge of migrating personalized soft prompts across foundation models by proposing PUMA, a lightweight adapter framework paired with an efficient user selection strategy. Our experiments on three large-scale datasets show that PUMA matches or surpasses the performance of full retraining while reducing computational costs by up to 98\%. The framework demonstrates strong generalization across diverse model architectures, robustness in complex chained migrations, and a novel capability to enhance personalization by fusing knowledge from multiple source models. By decoupling personalized assets from the underlying models, PUMA provides a practical and sustainable solution for the long-term evolution of personalized AI systems.

Our work opens several promising avenues for future research. One direction is to enhance the user selection process. While our group-based strategy is highly efficient, a learning-based approach, such as using Reinforcement Learning (RL) to train a selection policy, could further push the performance ceiling by discovering more potent user combinations than static heuristics.
Furthermore, our current framework focuses exclusively on migrating user-specific assets. We plan to extend PUMA to simultaneously migrate both user and item embeddings. This would enable a more comprehensive knowledge transfer, which is crucial in personalized systems where item representations are also learned and valuable.
Finally, we aim to address more complex and realistic migration scenarios, such as handling cold-start personalization. In cases where the target dataset contains new users not present in the source system, the learned migration adapter could be leveraged to quickly initialize their prompts. This would bypass the need for extensive training from scratch and offer an efficient solution for onboarding new users to the target model.

\section{Acknowledgments}
This work is supported by the National Natural Science Foundation of China (62402470),  the Fundamental Research Funds for the Central Universities of China (WK2100000053, PA2024GDSK0107), Anhui Provincial Natural Science Foundation (2408085QF189), the Postdoctoral Fellowship Program of CPSF (GZC20241643), and Anhui Postdoctoral Scientific Research Program Foundation (No.2025B1063). This research is supported by the advanced computing resources provided by the Supercomputing Center of the USTC.

\bibliography{aaai2026}
\end{document}